\documentclass{eptcs}
\providecommand{\event}{ICLP 2024} % This sets the conference name

\usepackage{iftex}
\usepackage{graphicx}
\usepackage{amsmath}
\usepackage{listings}
\usepackage{xcolor}
\usepackage{subcaption}
\usepackage{multirow}
\usepackage{url}

\ifpdf
  \usepackage{underscore}         % Only needed if you use pdflatex.
  \usepackage[T1]{fontenc}        % Recommended with pdflatex
\else
  \usepackage{breakurl}           % Not needed if you use pdflatex only.
\fi

\newcommand{\mypar}[1]{\paragraph{\bf #1.}}

\title{Logical Lease Litigation: Prolog and LLMs for Rental Law Compliance in New York}

\author{
    Sanskar Sehgal \qquad\qquad Yanhong A. Liu \\
    Stony Brook University, Stony Brook, NY 11794, USA \\
    \texttt{\{sasehgal, liu\}@cs.stonybrook.edu}
}
\def\titlerunning{Logical Lease Litigation}
\def\authorrunning{S. Sehgal \& Y. A. Liu}
\begin{document}

\maketitle

\begin{center}
    \rule{\textwidth}{0.5pt}\\
    \vspace{0.5em}
    \textbf{\large Abstract}\\
    \begin{abstract}Legal cases require careful logical reasoning following the laws, whereas
interactions with non-technical users must be in natural language.  As an application
combining logical reasoning using Prolog and natural language processing
using large language models (LLMs), this paper presents a novel approach
and system, LogicLease, to automate the analysis of landlord-tenant legal
cases in the state of New York.

LogicLease determines compliance with relevant legal requirements by
analyzing case descriptions and citing all relevant laws.  It leverages
LLMs for information extraction and Prolog for legal reasoning.  By
separating information extraction from legal reasoning, LogicLease achieves
greater transparency and control over the legal logic applied to each case.
We evaluate the accuracy, efficiency, and robustness of LogicLease through
a series of tests, achieving 100\% accuracy and an average processing time
of 2.57 seconds.  LogicLease presents advantages over state-of-the-art
LLM-based legal analysis systems by providing clear, step-by-step
reasoning, citing specific laws, and distinguishing itself by its ability
to avoid hallucinations---a common issue in LLMs.

   \end{abstract}
    \vspace{0.5em}
    \rule{\textwidth}{0.5pt}
\end{center}

% \begin{keywords}
% Legal case analysis, Prolog, Natural language processing, Large language models (LLMs), New York state law
% \end{keywords}

\section{Introduction}

Rental law compliance matters significantly to all rental residents. According to \cite{desilver2021national}, more than 122.8 million households in the United States are renters. \cite{hepburn2022preliminary} estimates that there are over 1.1 million cases of landlords evicting tenants every year, representing an increase of 75\% since 2021. Furthermore, 60\% of eviction case defendants in 2023 were women, and despite making up less than one-third of renters, nearly half of eviction case defendants in 2023 were Black. \cite{graetz2023comprehensive} states that over seven million Americans are evicted from their homes every year, nearly 40\% (2.7 to 3.2 million) of which are children. Additionally, in the United States, there is no guaranteed right to legal counsel in eviction proceedings. \cite{schultheis2019right} estimates that as many as 90\% of tenants facing eviction go to court unrepresented, putting them at a significant disadvantage. Consequently, up to 75\% of tenants end up losing their eviction cases.

The legal domain's emphasis on meticulous analysis and transparent thought processes makes it an ideal candidate for utilizing logic-based systems over black-box approaches. Logic-based systems excel in providing clear and transparent reasoning, which is highly valued in legal contexts. While large language models (LLMs) are increasingly being considered for automating legal analysis, they are prone to hallucinations, which can lead to incorrect legal interpretations.

This paper presents LogicLease, a novel system specifically designed to automate the analysis of landlord-tenant legal cases in the state of New York, providing a transparent and reliable alternative to black-box methods.

% write an overview of the problem, approach, and results

Rental law compliance is crucial for the well-being of tenants, but navigating legal cases can be complex and time-consuming. With millions of eviction cases each year, many tenants face eviction without proper legal representation, resulting in a significant disadvantage.

LogicLease harnesses the strengths of Large Language Models (LLMs) for information extraction and Prolog for legal reasoning. It is designed to assess compliance with relevant legal requirements by analyzing case descriptions presented in natural language. LogicLease utilizes LLMs to parse the input, extracting essential details that can influence case outcomes. Subsequently, it employs a logic-based backend to generate clear, step-by-step reasoning and cite specific laws. 

The accuracy, efficiency, and robustness of LogicLease were evaluated through a series of tests, demonstrating 100\% accuracy and an average processing time of 2.57 seconds. By separating information extraction from legal reasoning, LogicLease ensures greater transparency and control over the legal logic applied to each case. Furthermore, LogicLease addresses the challenges of hallucinations and opacity common in LLMs, providing a reliable tool for landlord-tenant legal analysis in New York.

The rest of the paper is organized as follows. Section~\ref{sec:problem} describes the problem of reliable analysis of rental law compliance. In Section~\ref{sec:approach}, we present the approach used in LogicLease for automating the analysis of landlord-tenant legal cases. Section~\ref{sec:implementation-evaluation} provides implementation details and evaluates the accuracy, efficiency, and robustness of LogicLease. Finally, Section~\ref{sec:related-conclusion} discusses related work and concludes the paper.

\section{The need for legal compliance analysis}
\label{sec:problem}

%describe the precise problem/requiremens: NL input, rigorous analysis, NL output

Millions of Americans face challenges in their homes, and those from marginalized communities are often disproportionately impacted by unfair housing practices. To address this gap, we developed LogicLease. LogicLease promises to provide free and highly accurate guidance on tenant-landlord issues. Our system is designed for simplicity, so anyone can navigate it, regardless of technical expertise. We prioritize accuracy to ensure users receive reliable information to confidently address their housing concerns.

When developing LogicLease, we established three primary requirements:
\begin{enumerate}
    \item Natural Language Input: Users can describe their situation in plain English (or potentially another language).
    This could involve describing the issue (e.g., repairs not being made, rent increase concerns), relevant details (lease terms, dates), and any questions they have.
    
    \item Rigorous Analysis for High Accuracy: The system uses LLMs to understand the user's situation and intent. It analyzes legal regulations, relevant case law, and best practices using the Prolog backend to provide accurate guidance. This might involve identifying the key legal issues involved, assessing the user's rights and responsibilities based on their location and lease agreement and considering potential solutions or next steps.
    
    \item Natural Language Output: The system provides clear and actionable information tailored to the user's situation. This output is generated using a Prolog backend, where description of relevant laws is coded as strings associated with the rules. As the Prolog backend processes the rules and retrieve the specific law applicable to the case, it generates and prints the corresponding description to the output. This may include explanations of tenant rights or step-by-step guidance on how to address issues (e.g., contacting the landlord, or filing a complaint), presented in a way that is easy to understand, even for people with no legal background.
\end{enumerate}

\section{Combining logic programming and LLMs for legal analysis}
\label{sec:approach}

% wirte a high-level design, try to be independent of "python" "SWI", "LLAMA" 
% (because any other language/tool with similar functions can do. 
% but Prolog is fine/good here:), and concrete examples from rental cases as examples are good too)
% the specific lang/tool details go to implementation section.

LogicLease consists of four main components:
\begin{enumerate}

\item Driver Script: This component serves as the central coordinator, orchestrating the interaction between the other components. It takes a case description (lease agreement) as input and invokes the Natural Language Processing (NLP) and Prolog functionalities for analysis.

\item Natural Language Processing (NLP): LogicLease utilizes a large language model (LLM) out-of-the-box, without any additional training, to extract attribute-value pairs from the lease agreement. These pairs represent critical aspects of the case, such as whether the lease is signed or the duration of the rental period.

\item Prolog Knowledge Base: This component contains a set of Prolog rules that represent legal requirements for rentals, which we manually coded based on the New York Renters' Rights Handbook~\cite{nytenantsrights2024}. It contains an exhaustive set of rules about lease validity, rent stabilization, eviction, habitability, and more. LogicLease utilizes the attribute-value pairs extracted by the LLM as arguments for Prolog queries and uses the Prolog engine to evaluate compliance against these rules.

\item User Interface: The user interface of LogicLease provides users with an intuitive and user-friendly experience, enabling them to interact with the system seamlessly (Figure~\ref{fig:fe1}). Users input the case description in natural language. Alternatively, they can also answer a series of dynamically generated questions related to the specific legal aspects of their case. These questions are designed to extract key attributes necessary for the legal analysis (Figure~\ref{fig:fe2}). Users can provide answers using dropdown menus or text input fields. The user interface processes the information and invokes the legal analysis.

\begin{figure}[h]
  \centering
  \begin{subfigure}{0.5\textwidth}
    \centering
    \includegraphics[width=\textwidth, height=0.9\textwidth, keepaspectratio]{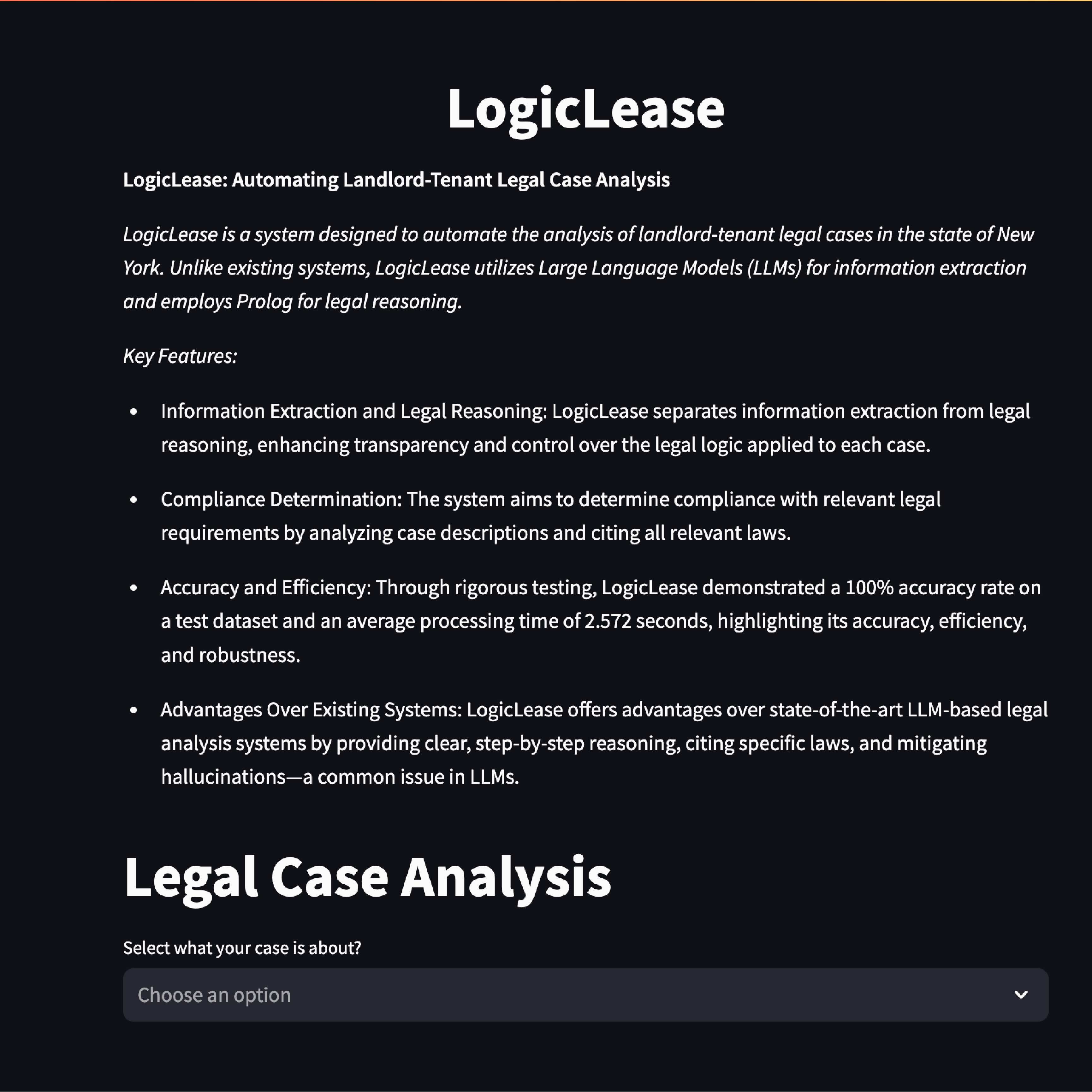}
    \caption{User Interface of LogicLease}
    \label{fig:fe1}
  \end{subfigure}%
  \begin{subfigure}{0.5\textwidth}
    \centering
    \includegraphics[width=\textwidth, height=0.9\textwidth, keepaspectratio]{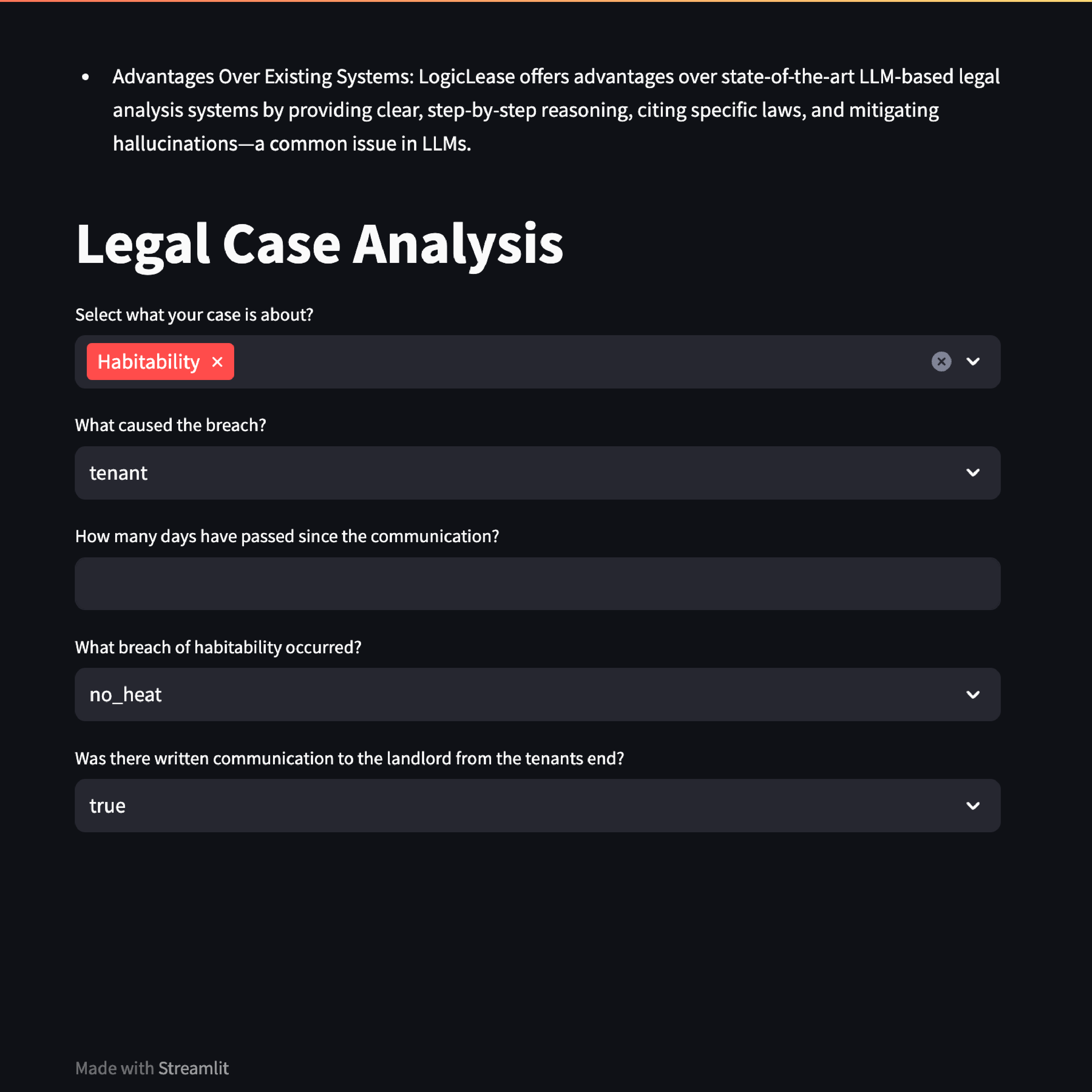}
    \caption{Example of dynamically generated questions}
    \label{fig:fe2}
  \end{subfigure}
  \caption{LogicLease front-end interface}
  \label{fig:frontend}
\end{figure}

\end{enumerate}

\begin{figure}[h]
  \centering
  \includegraphics[width=.9\textwidth, height=0.25\textheight]{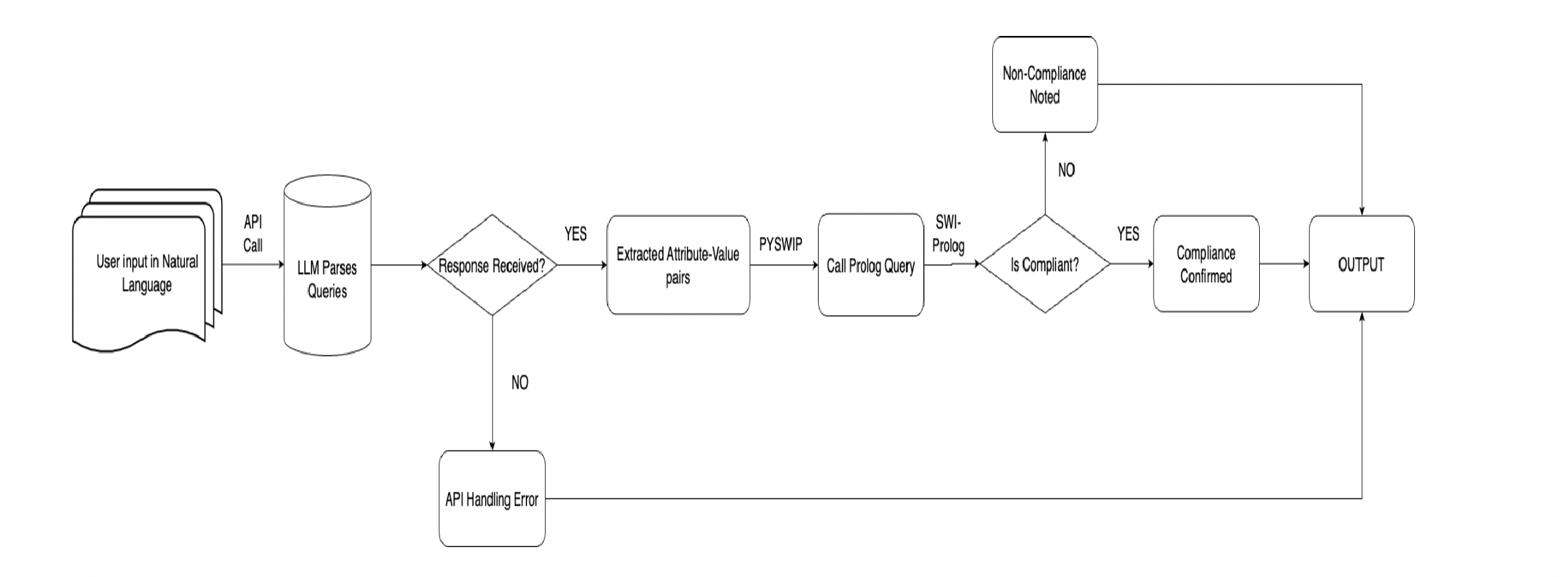}
  \caption{Design diagram of LogicLease}
  \label{fig:dataflow}
\end{figure}

As shown in Figure \ref{fig:dataflow}, the design of LogicLease follows a structured process. Initially, the Driver Script receives a case description in natural language as input. Subsequently, it calls the API to the LLM, providing a system query that specifies the desired lease agreement aspects and the user query containing the actual case description. The API call sends the queries to the LLM, retrieving a response containing extracted attribute-value pairs. The Driver Script parses this response and stores the extracted attribute-value pairs in a dictionary. Following this, the Driver Script invokes Prolog functions passing the dictionary containing the extracted attribute-value pairs. These functions translate the dictionary into a Prolog query based on the pre-defined Prolog rules in the knowledge base. The Prolog engine then evaluates the query, determining compliance with the lease agreement requirements. Finally, the results of the compliance check are displayed by the Driver Script.
\\

%make middel two boxes wider if needed, and others narrower. 
%the last box should be called output.
%mainly to make the words more regular font/readable.  
%Put the exceptional case on the bottom?

To illustrate the workflow of LogicLease and to facilitate a deeper understanding of the system, we provide an overview of the input processing, attribute-value pair extraction using an LLM, compliance checking with the Prolog backend system, and the final output.

\mypar{User input of case in natural language}
The input for a case is taken in natural language.  An example is the following:
\begin{quote}
\sf %or find some beter font
In a rent-stabilized apartment in Albany, New York, David, a disabled tenant, faced eviction proceedings initiated by his landlord, Ms. Johnson, citing owner occupancy as the cause. Despite David's disability, he has been asked to vacate. The matter has not been presented before court, and hence does not have a court ruling yet.
\end{quote}
This input is passed to the LLM for processing. 

\mypar{Attribute-value pairs extracted} %how about changing key to attribute everywhere?
The LLM parses the input text and outputs attribute-value pairs which are the core details of the case which influence the outcome. The extracted attribute-value pairs indicate that the eviction cause is "owner occupancy," there is no court ruling yet ("CourtRuling" is "false"), and the tenant is in a protected category ("TenantCategory" is "disabled"). These attribute-value pairs are stored in a dictionary in the following format. 

%don't need large braces, but need commas as separators?
\[
\begin{aligned}
&\text{EvictionCause}: \text{"owner\_occupancy"}, \\
&\text{CourtRuling}: \text{"false"}, \\
&\text{Executioner}: \text{"null"}, \\
&\text{TenantCategory}: \text{"disabled"},
\end{aligned}
\]
\\

\mypar{Logic rules and queries in Prolog}
As seen in Listing \ref{lst:eviction}, the compliance with legal requirements in this specific case is determined by invoking the following Prolog predicate. Notably, the arguments precisely match the information extracted by the Large Language Model (LLM) from the case description.

% use one with ttfont and with coloring

\begin{lstlisting}[language=Prolog, caption={Prolog rule for eviction compliance checking}, label={lst:eviction}, captionpos=b, basicstyle=\small\ttfamily, keywordstyle=\color{blue}, commentstyle=\color{green}, stringstyle=\color{red}]
eviction(EvictionCause, CourtRuling, Executioner, TenantCategory) :-
    eviction_law, 
    (CourtRuling == true ->
        (
            eviction_legal(EvictionCause),
            writeln('All conditions satisfied, eviction is lawful.'),
            nl
        )
    ;
    (
        eviction_warrant_execution(Executioner),
        \+ overrides(TenantCategory, EvictionCause),
        writeln('All conditions satisfied, eviction is lawful.'),
        nl
    )
).
\end{lstlisting}

%These extracted arguments are stored in a dictionary and utilized to invoke our Prolog predicate~\cite{wielemaker2012tplp,swi24}. 
The arguments of the predicate are instantiated based on the extracted attribute-value pairs stored in the dictionary.
The call to the Prolog predicate is structured as follows:
\begin{lstlisting}[language=Prolog, basicstyle=\small\ttfamily]
    Prolog.query(eviction(owner_occupancy, false, null, disabled).)
\end{lstlisting}

The Prolog query evaluates the compliance of the landlord's actions with the relevant legal requirements. 

\mypar{Final output to user in natural language}
The final output consists of a list of laws relevant to the case followed by a final judgment based on these laws. This transparent mechanism provides a clear rationale for each case.  For our example, the output is:

\begin{enumerate} %use the same/better font as for input
    \item \textbf{\sf Tenant with a lease is protected from eviction during the lease period if lease provisions and local laws are not violated.}
    \item \textbf{\sf Landlords must give formal notice before seeking legal possession of the apartment.}
    \item \textbf{\sf Eviction proceedings can be initiated for non-payment or significant lease violations.}
    \item \textbf{\sf Landlords of rent-regulated apartments may need DHCR approval for court proceedings.}
    \item \textbf{\sf Tenants should not ignore legal papers to avoid eviction.}
    \item \textbf{\sf Legal eviction requires court proceeding and judgment of possession.}
    \item \textbf{\sf Landlords cannot evict tenants unlawfully or by force.}
    \item \textbf{\sf Tenant evicted unlawfully can recover triple damages.}
    \item \textbf{\sf Additional rules protect certain groups from eviction.}
\end{enumerate}

\textcolor{red}{\textbf{\sf Tenant is in protected category, eviction for owner occupancy unlawful.}}
\\

The output confirms that the eviction for owner occupancy is unlawful given the tenant's protected category status and the absence of a court ruling.

\section{Implementation and evaluation}
\label{sec:implementation-evaluation}

%write an overview sentence
We have developed a complete implementation of the LogicLease system, covering the full New York State landlord-tenant legal framework. We successfully applied our system to a series of 10 test cases, representing various scenarios encountered in landlord-tenant disputes. To evaluate the system's performance, we employ a multi-pronged approach, assessing its accuracy, efficiency, and robustness across various dimensions.

\subsection{Implementation}

LogicLease is implemented using a combination of Python and SWI-Prolog ~\cite{wielemaker2012tplp,swi24}. Python is used to facilitate interactions among different components, utilizing libraries such as PYSWIP to establish communication between the Python script and the Prolog engine (SWIPL), while Streamlit is employed to create a user-friendly front-end interface. Additionally, llamaapi~\cite{llamaapi} is used to manage interactions with the Large Language Model (LLM) model LLaMA~\cite{touvron2023llama}, while Python libraries like json and ast are employed for processing the output received from LLaMA.

The Prolog backend serves as the foundation for legal reasoning, with custom clauses defining the relevant laws and procedures. The implementation of the system followed a structured approach, ensuring modularity and ease of maintenance. 

Notably, LogicLease incorporated defeasible logic~\cite{wan2009logic,wan2015defeasibility,morris24} within the Prolog component, enabling the system to handle situations where one legal principle takes precedence over another under specific circumstances.

By separating information extraction (NLP) from legal reasoning (Prolog), LogicLease achieved greater transparency and control over the legal reasoning applied to each case. This approach is particularly important for legal professionals who require a clear understanding of the system's reasoning process, while ensuring there are no hallucinations or snowballing effects.

The total size is approximately 500 lines of Prolog code and an additional 400 lines of Python code. 

All experiments and measurements were conducted on a macOS system featuring an Apple Silicon M2 processor, 8GB of RAM, and a 256GB SSD. The system was running macOS Monterey version 12.5, with Python 3.9.12, Prolog SWIPL 9.2.2, PYSWIP 0.2.11, llamaapi version 0.1.36, and Streamlit version 1.24.1.

\subsection{Accuracy}

To evaluate LogicLease's effectiveness, we manually compiled a dataset of lease litigation cases in New York. This dataset includes a mix of real-world cases (condensed for efficiency) and fictional scenarios we created. Due to limited API credits available for using the LLM, the system restricted the number of API calls made during the processing of legal cases. This limitation necessitated significant shortening of the text within the legal documents to fit within the allowed API usage. Despite these constraints, the resulting dataset of 10 cases effectively demonstrates the system's accuracy and usability in analyzing and responding to legal scenarios.

Encouragingly, in all ten cases tested, the LLM functioned effectively. It accurately extracted relevant details from each case description and successfully transferred this information to the Prolog backend system. The Prolog system, in turn, flawlessly interpreted the queries and delivered final verdicts on the legal issues presented. Although the dataset is currently small, the system's 100\% accuracy on this dataset helps build trust in the system.

To evaluate the accuracy of the system's reasoning, we conducted human-in-the-loop evaluation. This involved manually reviewing the system's output for each case. Legal resources such as handbooks and online legal forums (e.g., on Reddit) were used to verify the system's determinations. This process helped identify potential biases in the LLM's interpretation of the case details or errors within the Prolog code's reasoning logic.

\subsection{Efficiency}

We measured the average processing time per case, including the LLM extraction and the logic-based compliance check.  The processing times are shown in Table \ref{testcases}. The main bottleneck is the API call to the LLM Llama, which is not directly under our control.  However, solutions such as caching frequently accessed legal information can be explored.

While the LLM API call (to Llama) currently represents the performance bottleneck compared to Prolog's minimal processing time, the good news is that the average case processing time remains user-friendly at only 2.572 seconds.

It also helps if the case description passed to Llama is as concise as possible.  Not surprisingly, the time taken for Llama to respond is directly proportional to the length of the query string passed. In fact, the system exhibits a strong positive correlation (approx 68\%) between the length of the query string and Llama's response time.

\begin{table}[h]
    \centering
    \begin{tabular}{|c|c|c|}
        \hline
        \textbf{Test Case} & \textbf{Total Time} & \textbf{Prolog Running Time} \\
        \hline
        1 & 2.693 & 0.00012 \\
        2 & 2.462 & 0.00018 \\
        3 & 2.412 & 0.00021 \\
        4 & 3.156 & 0.00120 \\
        5 & 3.324 & 0.00130 \\
        6 & 2.270 & 0.00090 \\
        7 & 2.155 & 0.00050 \\
        8 & 2.215 & 0.00042 \\
        9 & 2.551 & 0.00031 \\
        10 & 2.486 & 0.00010 \\
        \hline
        \textbf{Average} & \textbf{2.572} & \textbf{0.00011} \\
        \hline
    \end{tabular}
    \caption{Running times, in seconds, for all test cases}
    \label{testcases}
\end{table}

%make the bars thinner
\begin{figure}[h]
  \centering
  \includegraphics[width=0.6\textwidth]{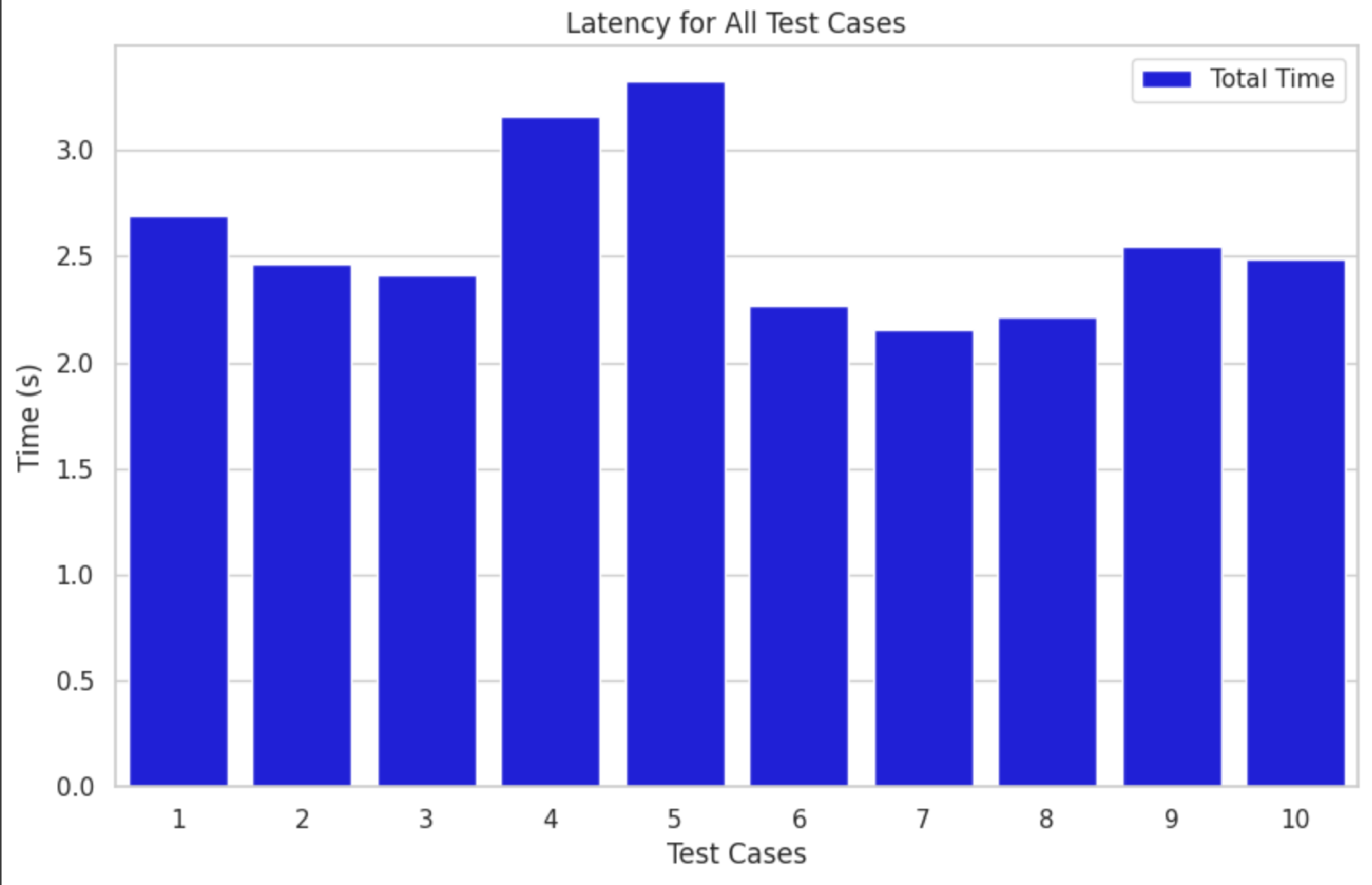}
  \caption{Total Time Taken for Each Test Case}
  \label{timetaken}
\end{figure}

\subsection{Robustness}

\begin{figure}[h]
  \centering
  \includegraphics[width=0.9\textwidth]{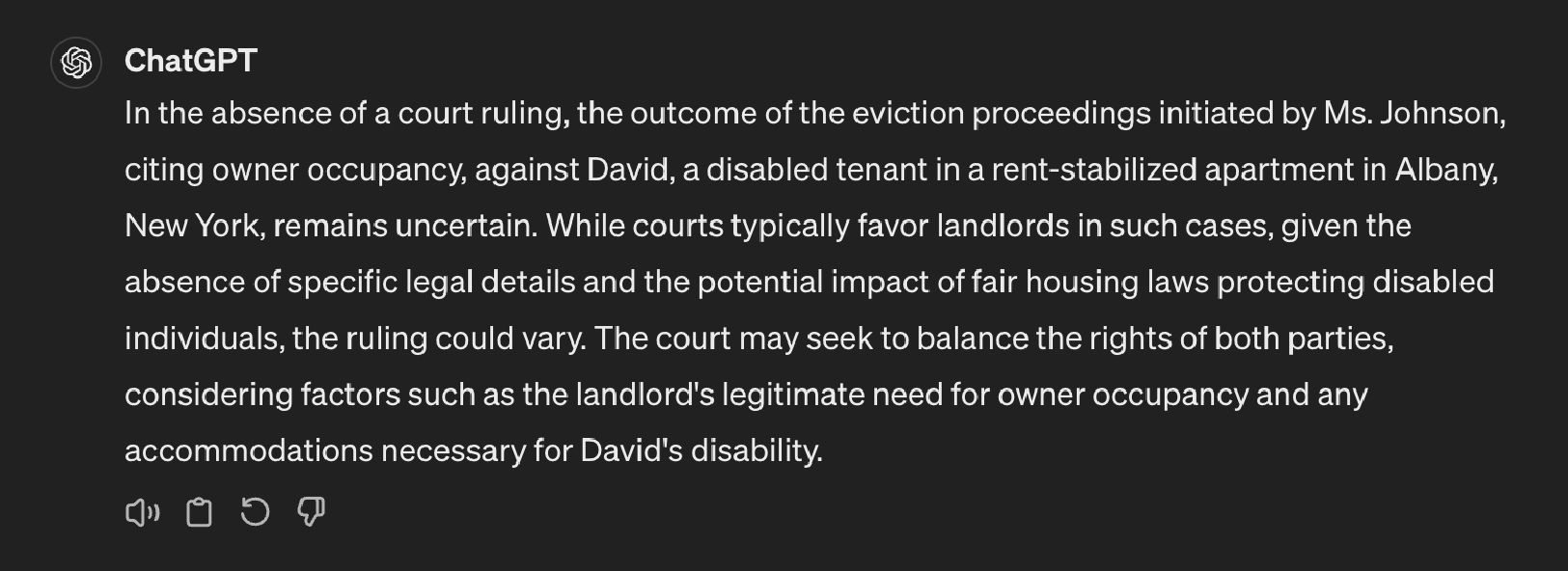}
  \caption{ChatGPT output}
  \label{chatgpt}
\end{figure}

\begin{figure}[h]
  \centering
  \includegraphics[width=0.9\textwidth]{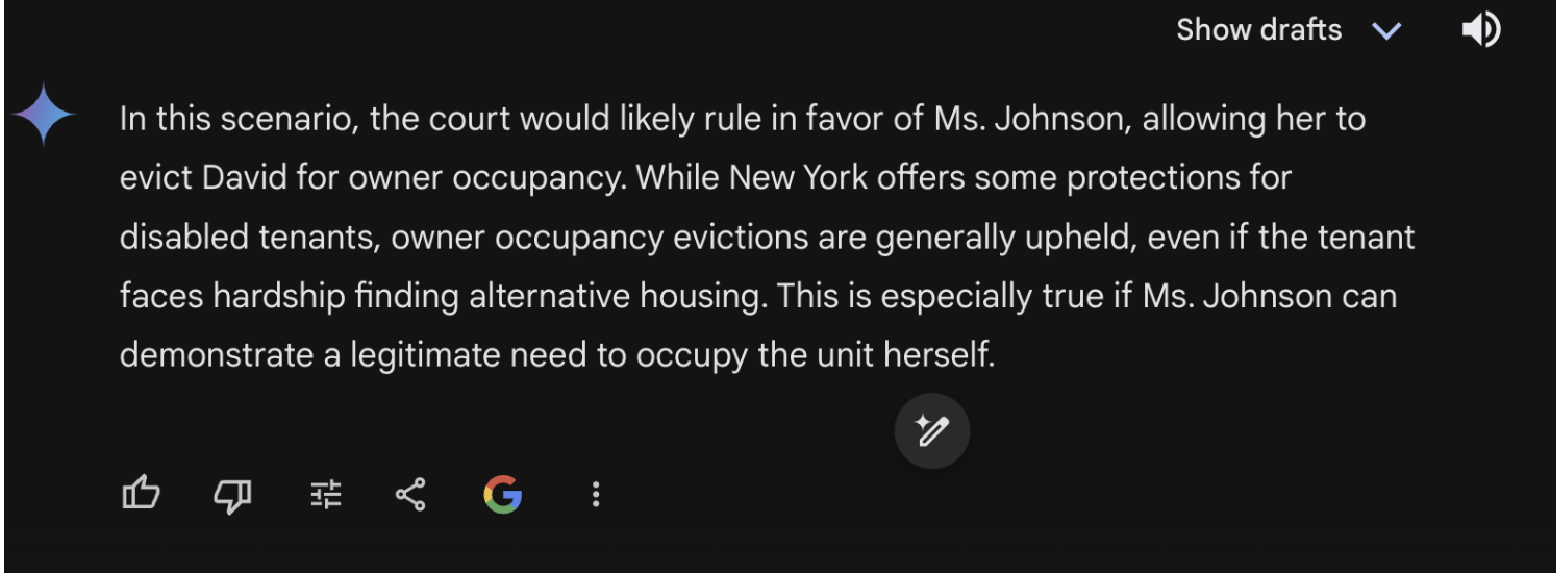}
  \caption{Gemini output}
  \label{gemini}
\end{figure}

To highlight the superiority and robustness of our system compared to existing solutions, we present a comparison with state-of-the-art Large Language Models (LLMs) such as ChatGPT \cite{chatgpt} and Gemini \cite{Gemini}.  These LLMs appear to favor the landlord in most cases, as shown by their outputs for the running example shown in Figures \ref{chatgpt} and \ref{gemini}.  Neither of these LLMs reached the correct conclusion for this example case. According to New York law, the situation clearly favors the tenant if they live in a rent-stabilized apartment and belong to a protected category.  For our test dataset, these LLMs did not reach the correct conclusion in three out of ten cases. This underscores the importance of our system and highlights the unreliability of LLMs for this application.

\section{Related work and conclusion}
\label{sec:related-conclusion}

Existing systems for legal analysis employ various techniques, including text classification, machine learning, and rule-based reasoning~\cite{avram2021pyeurovoc, lexnlp, kira, lawgeex, lexmachina}. However, these approaches often have limitations in handling nuanced legal reasoning. 

Text classification systems such as Legal-Document-Classifier \cite{avram2021pyeurovoc} and LexNLP \cite{lexnlp}  can categorize legal documents based on keywords and named entities but lack the ability to perform comprehensive legal reasoning. 

Machine learning-based systems like Kira \cite{kira} and LawGeex \cite{lawgeex} can extract key terms and identify potential issues in contracts, and models like Lex Machina \cite{lexmachina} can predict legal outcomes with some accuracy. However, these systems often operate as black boxes, raising concerns about transparency and fairness. They may also be limited by data quality and biases.

Rule-based legal reasoning systems like PROLEG~\cite{satoh2010proleg, prolegag} offer support for judges in civil litigation by incorporating predefined rules and handling uncertainty. However, their complexity can pose challenges for users.

In contrast, we successfully developed a system for analyzing landlord-tenant disputes in New York State by leveraging Large Language Models (LLMs) for information extraction and Prolog for legal reasoning. Achieving high accuracy and efficiency, the system offers several advantages over existing LLM-based legal analysis systems.

By separating information extraction from legal reasoning, the system achieves greater transparency and control over the legal logic applied to each case. Additionally, the use of Prolog enables the implementation of defeasible logic ~\cite{wan2009logic,wan2015defeasibility}, allowing the system to handle nuanced legal reasoning, such as resolving conflicting legal principles and dealing with uncertain or incomplete information~\cite{morris24}.

In conclusion, this work demonstrates the potential of combining Large Language Models (LLMs) and logic-based reasoning to create innovative tools for legal analysis. By addressing the limitations of existing approaches, LogicLease paves the way for more sophisticated and transparent systems in the field of legal technology.

Future work includes expanding the system's capabilities by employing techniques for caching frequently accessed legal information. Additionally, improving the testing process and adding more test cases will ensure better coverage and reliability of the system. Open-sourcing the code could encourage further development and broader adoption within the legal domain.

\nocite{*}
\bibliographystyle{eptcs}
\bibliography{generic}
\end{document}